%% file: acl_latex.tex
\pdfoutput=1
\documentclass[11pt]{article}

\usepackage[final]{acl}

\usepackage{times}
\usepackage{latexsym}
\usepackage[T1]{fontenc}
\usepackage[utf8]{inputenc}

\usepackage{microtype}

\usepackage{inconsolata}

\usepackage{graphicx}
\usepackage{hyperref}
\usepackage{url}
\usepackage{enumitem}
\usepackage[most]{tcolorbox}
\newcommand{\xhdr}[1]{{\noindent\bfseries #1}.}
\usepackage{booktabs}
\usepackage{booktabs}
\usepackage{colortbl}
\usepackage{caption}
\usepackage{wrapfig}
\usepackage{subfigure}
\usepackage{subcaption}
\usepackage{CJKutf8}
\usepackage{pifont}
\usepackage{amssymb} 
\usepackage{multirow}
\definecolor{mypink}{rgb}{.99,.91,.95}
\definecolor{mygreen}{rgb}{.9,.99,.9}
\definecolor{mygray}{gray}{.9}
\definecolor{Blue}{rgb}{0.05,0.05,0.4}
\definecolor{Red}{rgb}{0.4,0.05,0.05}

\title{DeepPrune: Parallel Scaling without Inter-trace Redundancy}

\author{ Shangqing Tu$^{1}$\thanks{~~Equal Contribution.},Yaxuan Li$^{2*}$, Yushi Bai$^1$,  Lei Hou$^1$\thanks{~~Corresponding author.}, Juanzi Li$^1$ \\
$^1$Tsinghua University, $^2$ShanghaiTech University \\
  \texttt{\{tsq25,bys22\}@mails.tsinghua.edu.cn} \\
  \texttt{liyx12023@shanghaitech.edu.cn,\{houlei,lijuanzi\}@tsinghua.edu.cn}  \\
\url{https://deepprune.github.io}
  }

\begin{document}
\maketitle

% \renewcommand{\thefootnote}{\fnsymbol{footnote}}
%     \footnotetext[2]{Equal contribution
%     }
% \renewcommand{\thefootnote}{\arabic{footnote}}

\input{000abstract}

\input{010intro}

\input{020preliminary}

\input{030method}

\input{040experiment}

\input{050conclusion}

\input{070limitation}

\input{080acknowledgement}

\bibliography{custom}

\input{060appendix}

\end{document}

%% file: 000abstract.tex
\begin{abstract}

Parallel scaling has emerged as a powerful paradigm to enhance reasoning capabilities in large language models (LLMs) by generating multiple Chain-of-Thought (CoT) traces simultaneously. However, this approach introduces significant computational inefficiency due to \textit{inter-trace redundancy}---our analysis reveals that over 80\% of parallel reasoning traces yield identical final answers, representing substantial wasted computation. To address this critical efficiency bottleneck, we propose \textbf{DeepPrune}, a novel framework that enables efficient parallel scaling through dynamic pruning. Our method features a specialized judge model trained with out-of-distribution data (AIME 2022, AIME 2023, and MATH 500) using oversampling techniques to accurately predict answer equivalence from partial reasoning traces, achieving 0.7072 AUROC on unseen reasoning models. Combined with an online greedy clustering algorithm that dynamically prunes redundant paths while preserving answer diversity. Comprehensive evaluations across three challenging benchmarks (AIME 2024, AIME 2025, and GPQA) and multiple reasoning models demonstrate that DeepPrune achieves remarkable token reduction of 65.73\%--88.50\% compared to conventional consensus sampling, while maintaining competitive accuracy within 3 percentage points. Our work establishes a new standard for efficient parallel reasoning, making high-performance reasoning more efficient. Our code and data are here: \url{https://deepprune.github.io/}.

\end{abstract}

%% file: 010intro.tex
\section{Introduction}

Large language models (LLMs)~\cite{GPT-4o,claude-3-5,reid2024gemini} have made remarkable progress on reasoning tasks~\cite{guo2025deepseek,team2025kimi,zeng2025glm}, especially when equipped with long Chain-of-Thoughts (CoT) that can mimic human's thinking processes~\cite{wei2022chain,sprague2024cot}. This advancement is driven by inference-time scaling~\cite{jaech2024openai}, a new paradigm that enhances LLM's reasoning capabilities via more computing in the test stage~\cite{snell2025scaling}.  

\begin{figure}[t]
    \centering
    \includegraphics[width=\linewidth]{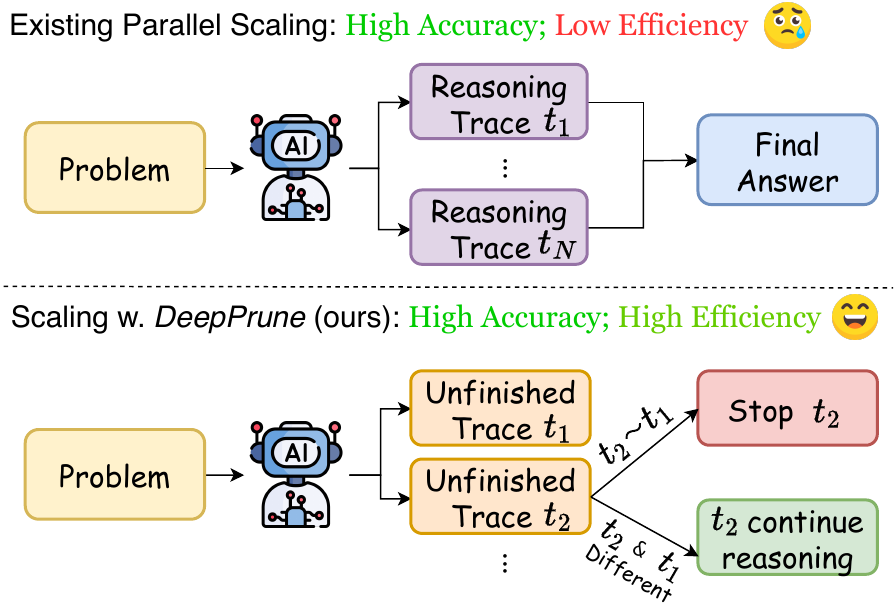}
    \caption{\textit{DeepPrune} conducts early stopping based on the similarity between reasoning treaces to enhance the efficiency of parallel scaling and save diverse traces.}
    \label{fig:fig1_intro}
\end{figure}

Generally, there are two types of inference-time scaling: sequential scaling and parallel scaling~\cite{venkatraman2025recursive}. Sequential scaling~\cite{muennighoff2025s1} focuses on increasing the computation in one reasoning trace like expanding the output length to 128k. While parallel scaling (e.g. best-of-n sampling) encourages generating multiple reasoning traces simultaneously, further pushing the total token cost to 100M or higher~\cite{moshkov2025aimo}. However, beneath these advances lies a practical question: \textbf{How to achieve high performance with low token cost?}
% includes: best-of-n, majority voting, GenSelect, etc.,

Existing efficient reasoning methods mainly focus on alleviating the over-thinking of sequential scaling~\cite{chen2024not,hou2025thinkprune,zhang2025adaptthink}. There are few works designed for parallel scaling~\cite{madaan2025rethinking}, which typically adopt the LLM's internal signal like confidence~\cite{fu2025deep} for early stopping to improve the sampling efficiency. However, these confidence-based methods suffer from two fundamental limitations: (1) they fail to reduce redundancy \textit{between} parallel reasoning paths, and (2) they risk prematurely terminating correct reasoning traces.

In this paper, we propose \textbf{DeepPrune}, as shown in Figure~\ref{fig:fig1_intro}, a novel method that proactively prunes redundant parallel CoTs while preserving traces with diverse answers. Our approach is motivated by a key observation from preliminary experiments: approximately 80\% of parallel reasoning traces yield identical final answers, while only 20\% produce distinct solutions. This reveals significant redundancy in current parallel reasoning paradigms.

We further investigate whether early-stage trace similarity can predict final answer equivalence. Surprisingly, shallow semantic similarity measures (e.g., SentenceBERT on first 500 tokens) achieve only random-level performance (AUROC=0.58), while deeper LLM-based comparison (\texttt{Qwen3-4B-Instruct}) shows moderate improvement (AUROC=0.66) but remains suboptimal for practical deployment. This finding underscores the necessity for specialized models capable of understanding reasoning processes at a deeper level.

Inspired by this analysis, we train a LLM-based judge model that predicts redundancy between truncated reasoning traces. To enable accurate early stopping, we explore two truncation strategies including fixed-length prefixes and reasoning-step aligned segments. To address class imbalance and preserve answer diversity, we employ focal loss and oversampling techniques for training the judge model. For efficient online inference, we design a greedy clustering algorithm that dynamically prunes redundant paths during generation.

We conduct comprehensive experiments to prove the effectiveness of DeepPrune across diverse settings. To ensure robust generalization, we train our judge model on fully out-of-distribution data (AIME 2022, AIME 2023, and MATH 500) and evaluate it on unseen reasoning models and benchmarks. In offline evaluation, our judge model achieves an average AUROC of 0.7072 and TPR of 0.5063 when using first-800 tokens with oversampling, demonstrating strong cross-model generalization to three unseen reasoning models. More importantly, in online reasoning tasks across three challenging benchmarks (AIME 2024, AIME 2025, and GPQA) and three state-of-the-art reasoning models (DeepSeek-8B, Qwen3-32B, and GPT-OSS-20B), DeepPrune reduces token consumption by 65.73\% to 88.50\% compared to cons@512 which samples 512 traces and conducts majority voting, while maintaining comparable accuracy (within 3.4 percentage points). Notably, on AIME24 and AIME25 datasets, DeepPrune achieves up to 88.5\% token reduction while maintaining or improving accuracy, substantially outperforming strong baselines like DeepConf~\cite{fu2025deep}.

Our contributions are threefold: (1) We identify and quantify the pervasive problem of \textit{inter-trace redundancy} in parallel reasoning, revealing that over 80\% of computational resources are wasted on generating equivalent reasoning paths. (2) We propose DeepPrune, a novel framework that combines a trained judge model with online greedy clustering to efficiently prune redundant reasoning traces while preserving answer diversity. (3) Extensive offline and online experiments show that our method reduces token consumption by up to 88.5\% without compromising accuracy, significantly outperforming existing baselines across multiple reasoning benchmarks and model architectures.

%% file: 020preliminary.tex
\section{Related Work}

\begin{figure*}[htbp]
    \centering
    \subfigure[Redundant Traces Distribution]{
        \includegraphics[width=0.31\linewidth]{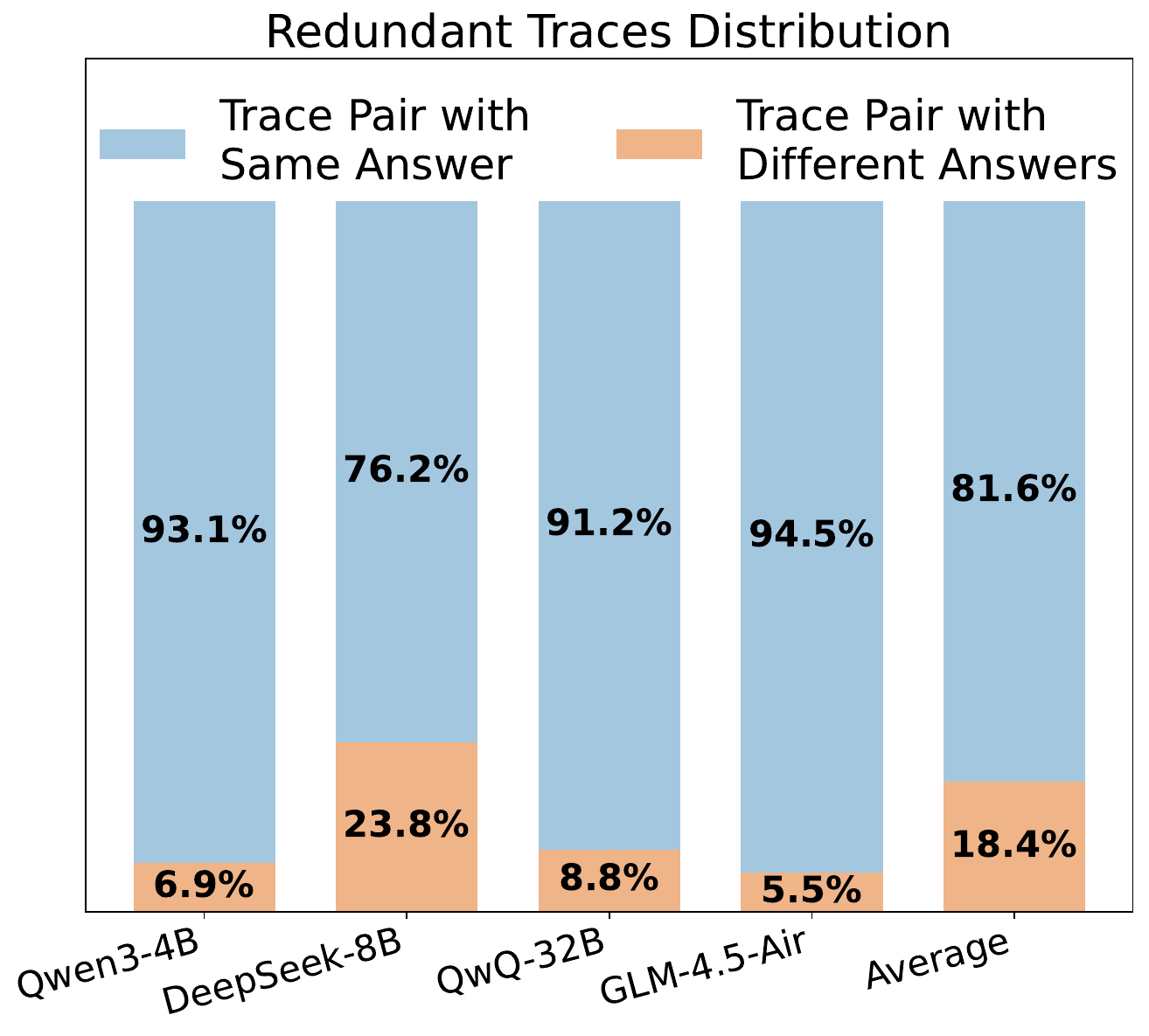}
        \label{fig:subfig1}
    }
    \hfill
    \subfigure[ROC of Semantic Similarity]{
        \includegraphics[width=0.32\linewidth]{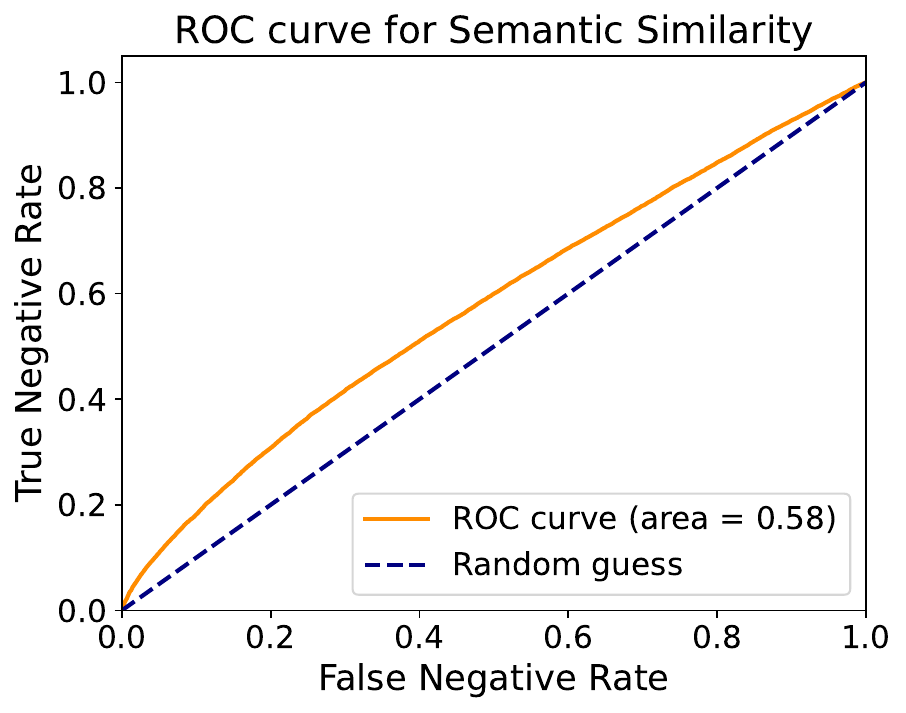}
        \label{fig:subfig2}
    }
    \hfill
    \subfigure[ROC of LLM Judge]{
        \includegraphics[width=0.32\linewidth]{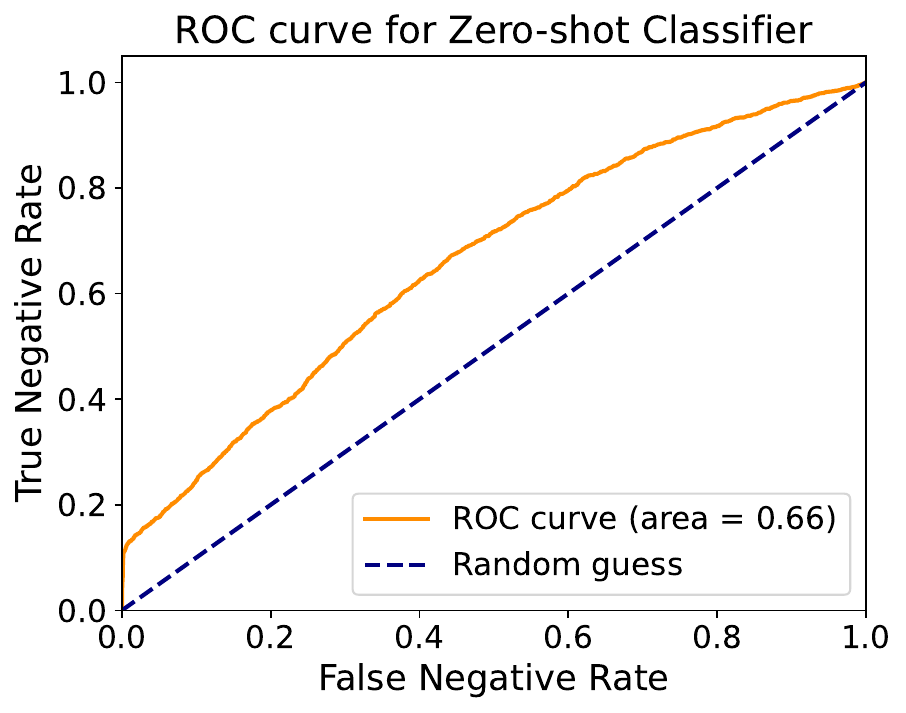}
        \label{fig:subfig3}
    }
    \caption{Analysis of Inter-trace Redundancy. (a) Distribution of same vs. different answer pairs of reasoning traces, revealing severe redundancy. (b) ROC curve for shallow semantic similarity (SentenceBERT) to distinguish traces with same answers from those with different ones, which shows limited predictive power (AUROC=0.58). (c) ROC curve for LLM-based deep comparison (Qwen3-4B-Instruct) achieves moderate improvement (AUROC=0.66).} %but remains suboptimal, underscoring the necessity for a more effective prediction mechanism to enable early pruning of redundant reasoning paths
    \label{fig:three-subfigs}
\end{figure*}
\xhdr{Parallel Scaling} Parallel scaling has emerged as a pivotal paradigm for enhancing reasoning performance through concurrent generation of multiple reasoning traces~\cite{chen2024more,pan2025learning,zheng2025parallel}. Self-Consistency~\cite{wang2022self} pioneered majority voting over diverse reasoning paths, while Best-of-$N$ sampling~\cite{brown2024large} extended this concept through explicit candidate ranking. 
There are also tree-based exploration methods like ToT~\cite{yao2023tree} that dynamically branch reasoning paths. 

\xhdr{Efficient Reasoning}
Efficient reasoning methods~\cite{feng2025efficient,sui2025stop,liu2025efficient,qu2025survey} aim to optimize the accuracy-compute trade-off during inference~\cite{kang2025c3ot,srivastava2025towards,li2025adaptive}. Prior research has explored reducing token usage in individual reasoning traces, such as through length-conscious fine-tuning~\cite{liu2024can,arora2025training,aggarwal2025l1,xia2025tokenskip} or training-free prompting techniques~\cite{renze2024benefits,han2024token,xu2025chain,fu2025reasoning,aytes2025sketch}. Another line of work improves parallel scaling efficiency by early-stopping redundant samples via confidence estimates~\cite{fu2025deep,yang2025dynamic} or by refining aggregation strategies~\cite{wang2024soft,wang2025ranked}. While these methods address intra-trace verbosity or sample quantity reduction, they do not explicitly model redundancy \textit{between} parallel reasoning paths. Our work directly targets this inter-trace redundancy, enabling proactive pruning while preserving answer diversity.

% TODO, 按论文里的定义，1是positive, 是相同，0是不同。所以这里x轴改成FNR，y轴改成TNR就行

\section{Preliminaries}
\subsection{Problem Definition}
\label{sec:problem_def}

Given a set of $n$ parallel reasoning traces $S_1 = \{t_1, t_2, ..., t_n\}$ generated concurrently for the same query, our objective is to reduce inter-trace redundancy while preserving answer diversity. We define a pruning process $P$ that selects a subset of traces:

\[
P(S_1) = S_2,\quad \text{where } S_2 \subseteq S_1
\]

The pruned set $S_2$ should satisfy:

\[
S_2 = \left\{ t_{k_1}, t_{k_2}, \dots, t_{k_m} \;\middle|\; 
\begin{aligned}
&\text{sim}(t_{k_i}, t_{k_j}) < \tau,\\
&\forall i,j \leq m
\end{aligned}
\right\}
\]
where $\text{sim}(t_{k_i}, t_{k_j}) $ is the similarity, $\tau$ is a similarity threshold. The traces in $S_2$ continue reasoning to produce final answers $\{o_{k_1}, o_{k_2}, \dots, o_{k_m}\}$.

To operate this process, we need to model the \textit{Similarity} function. Since parallel reasoning focuses on answer accuracy, we simplify the judgment to predicting whether two incomplete traces will yield the same final answer. Formally, for any pair of unfinished traces $(t_i, t_j)$, we predict whether their corresponding results $(o_i, o_j)$ will be identical, which can be defined as the binary similarity function based on final answer equivalence:
\[
\text{sim}(t_i, t_j) = 
\begin{cases} 
1 & \text{if } R(o_i, o_j) = 1 \\
0 & \text{if } R(o_i, o_j) = 0 
\end{cases}
\]
where $R(o_i, o_j)$ is the reward function for answer equivalence based on verifable rules or reward models. In this paper, we only consider queries with verifable answers, leaving others for future works.
% 引用RLVR等

\begin{figure*}[t]
    \centering
    \includegraphics[width=\linewidth]{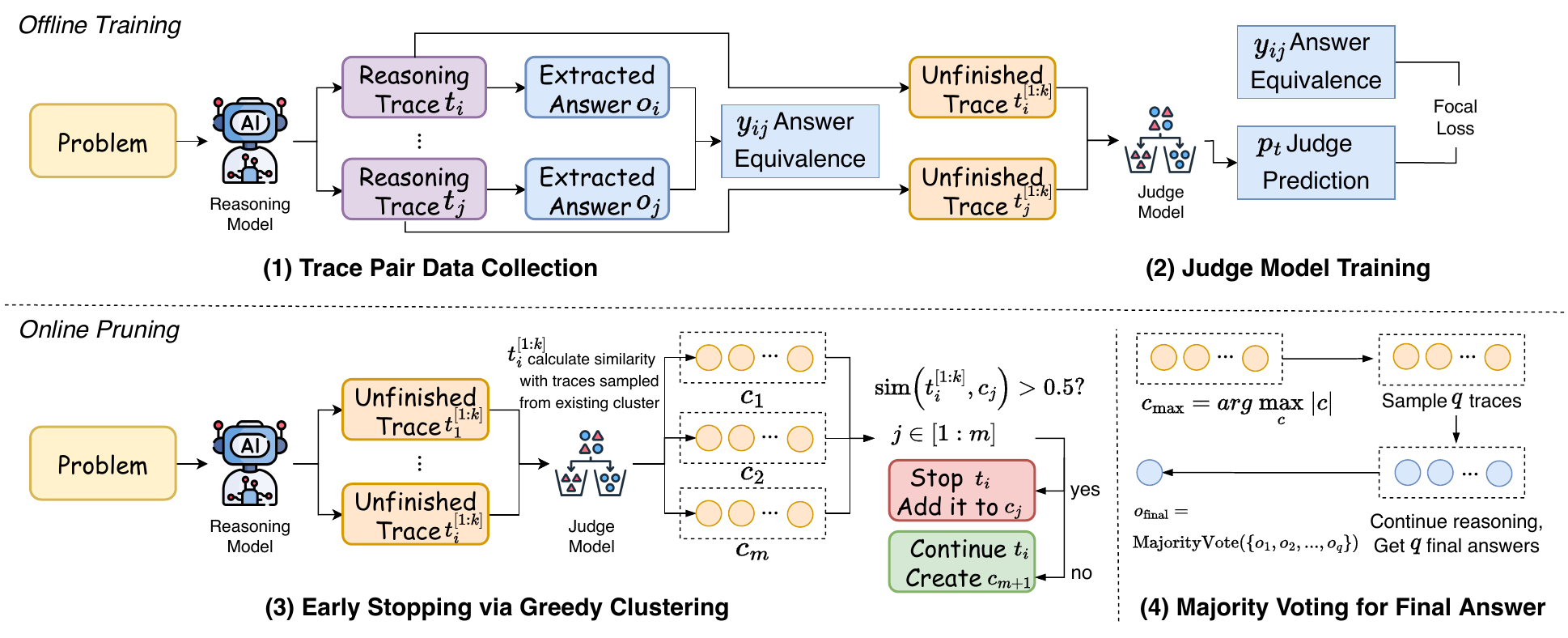}
    \caption{Overview of the \textbf{DeepPrune} framework. The \textit{offline training} phase (top) involves constructing trace pair datasets with binary labels indicating answer equivalence, then training a judge model using focal loss and oversampling to address class imbalance. The \textit{online pruning} phase (bottom) leverages the trained judge model to perform dynamic pruning via greedy clustering where traces are assigned to existing clusters or new ones based on similarity predictions, and concludes with majority voting on selected traces to determine the final answer.}
    %\vspace{-2mm}
    \label{fig:pipeline}
\end{figure*}

\subsection{Inter-trace Redundancy: The Efficiency Bottleneck of Parallel Reasoning}
\label{sec:pre_exp}
Recent advances in parallel scaling have significantly improved reasoning performance, 
but at the cost of substantial computational overhead. 
We identify \textbf{inter-trace redundancy} as the primary efficiency bottleneck: 
when generating multiple reasoning traces in parallel, a large proportion of tokens 
are wasted on producing semantically equivalent reasoning paths that lead to identical answers.

\xhdr{Parallel Reasoning Trace Collection}
To uncover the phenomenon of inter-trace redundancy, we conduct a reasoning trace collection process. We select four widely-used reasoning models:  \texttt{Deepseek-R1-Distill-Llama-8B}, \texttt{Qwen3-4B-Thinking-2507}, \texttt{GLM-4.5-Air} and \texttt{QWQ-32B} . These models are evaluated on over 100 problems from reasoning benchmarks including Math500~\cite{hendrycksmath2021}, AIME24, AIME25~\cite{aime25}, and GPQA~\cite{rein2024gpqa}. For each problem, we generate 16 parallel reasoning traces per model. The traces for each problem are paired exhaustively, resulting in $\binom{16}{2} = 120$ pairs per problem. Thus, for each model, we obtain around 12,000 pairs in total. To determine answer equivalence, we use the rule-based reward function from DeepScaleR~\cite{deepscaler2025} to verify whether the final answers of each pair are identical. More details are in Appendix~\ref{app:data_detail}.

The results, summarized in Figure~\ref{fig:subfig1}, reveal a striking dominance of same-answer pairs across all models, with ratios exceeding 80\% in most cases. Specifically, \texttt{GLM-4.5-Air}~\cite{zeng2025glm} has 94.5\% same-answer pairs. This severe class imbalance highlights that even with a modest number of samples (16 per problem), a large proportion of computational resources are wasted on generating redundant reasoning paths. The high prevalence of same-answer pairs (over 80\% on average) underscores inter-trace redundancy as a critical efficiency bottleneck in parallel reasoning.

\xhdr{Preliminary Experiment}
We next investigate whether the similarity between unfinished traces can predict the similarity of their final answers. This capability is crucial for early pruning of redundant paths. We evaluate two approaches for this prediction task: (1) \textit{Shallow Semantic Similarity:} First, we use SentenceBERT (\texttt{all-MiniLM-L6-v2} model~\cite{reimers2019sentence}) to compute cosine similarity between the first 700 tokens of two traces. This similarity score serve as a feature for binary classification. As shown in Figure~\ref{fig:subfig2}, the ROC curve achieves an AUROC of only 0.58, which is barely better than random guessing (AUROC=0.5). This indicates that surface-level semantic features are insufficient for predicting answer equivalence. (2) \textit{LLM-based Deep Comparison:} To leverage deeper  understanding for reasoning traces, we employ \texttt{Qwen3-4B-Instruct} in a zero-shot setting. We design a prompt that instructs the model to compare two unfinished traces with 700 tokens and judge whether they will yield the same final answer. The ROC curve for this classifier achieves an AUROC of 0.68, a notable improvement over SentenceBERT. However, this performance remains suboptimal for practical deployment.

These results demonstrate that while LLM-based judgment captures logical equivalence better than semantic similarity, there is significant room for improvement. The suboptimal performance of both approaches motivates our proposed method.

%% file: 030method.tex
\section{DeepPrune}
\label{sec:method}

To address the inter-trace redundancy problem in parallel scaling, we propose \textbf{DeepPrune}, a two-stage framework that includes offline training of a specialized judge model and online inference-time pruning. As demonstrated in Figure~\ref{fig:pipeline}, the core idea is that by accurately predicting whether two incomplete reasoning traces will yield identical final answers, we can efficiently prune redundant paths while preserving answer diversity.

\subsection{Offline Training}
\label{sec:sft_data_collection}
\subsubsection{Trace Pair Data Collection}
\label{sec:data_collection}

To train our judge model, we construct a dataset of reasoning trace pairs with binary labels indicating whether they lead to identical final answers. For each input query q, we generate n parallel reasoning traces $\{t_1, t_2, \dots, t_n\}$ using the same reasoning model. The traces are paired exhaustively, resulting in $\binom{n}{2}$ pairs per query. The similarity label $y_{ij}$ of each pair $(t_i, t_j)$ is based on answer equivalence:

\[
y_{ij} = R(o_i, o_j)
\]
where $R(o_i, o_j)$ is a reward function that verifies answer equivalence using rule-based methods from DeepScaler, and $o_i, o_j$ are the final answers derived from traces $t_i$ and $t_j$ respectively.

A key challenge is determining how to extract meaningful segments from unfinished traces for early redundancy prediction. We explore two truncation strategies: (1) \textbf{Fixed-length prefix:} Truncate the first k tokens from each trace: $t_i^{[1:k]}$ and $t_j^{[1:k]}$. (2) \textbf{Reasoning-step alignment:} Extract segments containing the same number of reasoning steps, which can be represented by first $k$ reasoning words like \textit{wait}, \textit{thus}, and \textit{since} that drive the direction of reasoning pathways~\cite{wang2025beyond}.

Our training data is collected exclusively from \texttt{Deepseek-R1-Distill-Llama-8B} outputs using out-of-distribution datasets (AIME 2022, AIME 2023, and MATH 500), while traces from other models and different datasets are reserved for testing cross-model generalization.

\subsubsection{Judge Model Training Strategy}
\label{sec:judge_training}

We fine-tune \texttt{Qwen3-4B-Instruct} as our generative judge model $J_\theta$ to predict the similarity label $y_{ij}$ given a pair of unfinished traces $(t_i, t_j)$. The model takes the concatenated trace pair as input and outputs a binary prediction:

\[
\hat{y}_{ij} = J_\theta(\text{concat}(t_i, t_j))
\]

To address the severe class imbalance where same-answer pairs constitute approximately 80\% of the data, we employ two complementary techniques:

\xhdr{Focal Loss}
We use focal loss~\cite{lin2017focal} to focus training on hard negative examples (different-answer pairs), improving the true negative rate:

\[
\mathcal{L}_{focal} = -\alpha_t(1-p_t)^\gamma \log(p_t)
\]
where $p_t$ is the model's estimated probability for the true class, $\gamma$ modulates the rate at which easy examples are down-weighted, and $\alpha_t$ balances class importance for the 80/20 distribution of our labels.

\xhdr{Oversampling}
We oversample the minority class by a factor of 2 to achieve balanced class distribution during training, ensuring the model receives sufficient exposure to diverse reasoning patterns.

\subsection{Online Pruning}
\label{sec:online_reasoning}

\subsubsection{Early Stopping via Greedy Clustering}
\label{sec:early_stopping}

During online pruning, we generate multiple parallel traces simultaneously and dynamically prune redundant paths. Let $S = \{t_1, t_2, \dots, t_N\}$ be the set of $N$ parallel reasoning traces. Our goal is to select a diverse subset $S' \subseteq S$ that maximizes answer diversity while minimizing computational cost.

We propose a greedy clustering algorithm that operates with unfinished traces. The algorithm maintains a set of clusters $C = \{c_1, c_2, \dots, c_m\}$, where each cluster represents traces that are predicted to yield identical answers $c_j = \{t_{1}, t_{2}, \dots, t_{|c_j|}\}$. For each new trace $t_i \in S$, we compute its average similarity with representative traces sampled from the existing cluster $c_j \in C$:

\[
\text{sim}(t_i, c_j) = \frac{1}{p} \sum_{h=1}^{p} J_\theta(t_i, t_h^{(j)})
\]
where $t_h^{(j)}$ are randomly sampled top-$p$ traces from cluster $c_j$ with $p = \min(20, |c_j|)$. If $\max_j \text{sim}(t_i, c_j) > \tau$, we assign $t_i$ to the most similar cluster $\arg\max_j \text{sim}(t_i, c_j) $; otherwise, we create a new cluster if the maximum number of clusters $K$ has not been reached. If $K$ is reached, we will terminate the clustering process.

\begin{table*}
\centering
\resizebox{\linewidth}{!}{
\begin{tabular}{l|c|c|c|c|c|c|c|c}
\toprule
\multirow{2}{*}{\textbf{Judge Training Method}} & \multicolumn{2}{c}{\textbf{Average}} & \multicolumn{2}{|c}{\textbf{Qwen3-4B-Thinking}} & \multicolumn{2}{|c}{\textbf{QwQ-32B}} & \multicolumn{2}{|c}{\textbf{GLM-4.5-Air}} \\
 & \multicolumn{1}{|c}{\textbf{AUROC}} & \textbf{TNR@0.2} & \multicolumn{1}{|c}{\textbf{AUROC}} & \textbf{TNR@0.2} & \multicolumn{1}{|c}{\textbf{AUROC}}  & \textbf{TNR@0.2} & \multicolumn{1}{|c}{\textbf{AUROC}} & \textbf{TNR@0.2} \\
\midrule
Top-500 Tokens & 0.8556 & 0.7720 & 0.8582 & 0.7720 & 0.8435 & 0.7632 & 0.8652 & 0.7808 \\
 + Focal loss & 0.8360 & 0.7327 & 0.8369 & 0.7134 & 0.8309 & 0.7373 & 0.8401 & 0.7473 \\
 + Oversampling & 0.7610 & 0.5232 & 0.7587 & 0.4910 & 0.7509 & 0.5154 & 0.7733 & 0.5632 \\
+ Focal loss \& Oversampling & 0.8608 & 0.7698 & 0.8710 & 0.7869 & \textbf{0.8586} & 0.7629 & 0.8528 & 0.7595 \\
\midrule
Top-25 Reasoning Words & 0.8326 & 0.6647 & 0.7948 & 0.5236 & 0.8190 & 0.6587 & 0.8841 & 0.8117 \\
+ Focal loss & 0.8559 & 0.7403 & 0.8434 & 0.6846 & 0.8253 & 0.6842 & \textbf{0.8989} & 0.8522 \\
+ Oversampling & 0.7983 & 0.6762 & 0.8095 & 0.6677 & 0.7917 & 0.6521 & 0.7938 & 0.7089 \\
+ Focal loss \& Oversampling & \textbf{0.8701} & \textbf{0.8186} & \textbf{0.8705} & \textbf{0.8100} & 0.8512 & \textbf{0.7905} & 0.8886 & \textbf{0.8554} \\
\bottomrule
\end{tabular}
}
\caption{The offline evaluation results of the judge model across different truncation methods and training strategies, which reports the average AUROC and TNR@0.2 metrics for three reasoning models using two truncation types: top-500 tokens and top-25 reasoning words, with the combination of focal loss and oversampling. }
\label{tb:method_comparison}
\end{table*}

Our approach reduces the number of similarity judgments compared to exhaustive pairwise comparisons, making it suitable for real-time inference.

\subsubsection{Majority Voting for Final Answer}
\label{sec:majority_voting}

After clustering, we need to select a final answer from the remaining traces. Since our judge model may give wrong prediction, we observe two kinds of errors: (1) Most pairs are classified as equivalent, so the largest cluster has too many traces. (2) All trace pairs are predicted as different, therefore each cluster only has one trace. To deal with these situations, we first select the largest cluster $c_{\text{max}}$ ($|c|$ means the number of reasoning traces in $c$):

\[
c_{\text{max}} = \arg\max_{c \in C}  |c|
\]

To conduct voting without too many identical traces, we only let top-$q_1$ traces in $c_{\text{max}}$ to finish reasoning, where $q_1 = \min( |c_{\text{max}}|, 20)$. Besides, if all clusters are singletons, i.e. $|c| = 1, \forall c \in C$, which means the judge model is highly likely wrong, we just sample $q_2 = 32$ traces from $S$ for final reasoning. Finally, we apply majority voting on the final answers of those finished traces:

\[
o_{\text{final}} = \text{MajorityVote}(\{o_1, o_2, \dots, o_q\})
\]
where $q = q_1$ or $q = q_2$ depending on situations.

This approach ensures that we can invest computational resources primarily in promising reasoning paths even if the judge model may produce wrong prediction, which reduces token consumption in parallel reasoning while preserving answer quality.

% The overall DeepPrune framework demonstrates that by combining learned similarity judgment with efficient online pruning, we can significantly reduce token consumption in parallel reasoning while preserving answer quality.

%% file: 040experiment.tex
\section{Experiments}

\subsection{Experimental Setup}

\xhdr{Settings} Our evaluation includes both offline assessment of the judge model's predictive performance and online testing of the full pruning framework during reasoning tasks. For offline evaluation, we train our judge model on reasoning traces from \texttt{Deepseek-R1-Distill-Llama-8B} and evaluate its generalization capability on three distinct reasoning models: \texttt{Qwen3-4B-Thinking-2507}, \texttt{QWQ-32B}, and \texttt{glm-4.5-air}. We use reasoning traces collected from challenging reasoning benchmarks comprising over 1,000 problems. For online reasoning experiments, we evaluate three reasoning models: \texttt{DeepSeek-8B}~\cite{guo2025deepseek}, \texttt{Qwen3-32B}~\cite{yang2025qwen3}, and \texttt{GPT-OSS-20B}~\cite{agarwal2025gpt} across the three benchmarks: AIME 2024~\cite{aime24}, AIME 2025~\cite{aime25}, and GPQA~\cite{rein2024gpqa}. We generate 512 parallel reasoning traces per problem for baseline methods and apply DeepPrune with a redundancy threshold $\tau = 0.5$. 
% All experiments are conducted on 4 NVIDIA A100 GPUs with 80GB memory.

\xhdr{Metrics}
We employ two categories of evaluation metrics. (1) \textbf{Offline Evaluation Metrics:} We assess the judge model's binary classification performance using AUROC (area under the receiver operating characteristic curve) to measure overall classification performance, and TNR@0.2 (true negative rate at false negative rate of 0.2) to evaluate the model's ability to identify diverse reasoning paths while controlling false negatives. (2) \textbf{Online Evaluation Metrics:} We measure the end-to-end system performance using token consumption, accuracy (final answer correctness measured by exact match with ground truth), and token reduction percentage reduction in token consumption compared to original consensus sampling (cons@512):
\[
\Delta\text{Token}\% = \frac{\text{Tokens}_{\text{new}} - \text{Tokens}_{\text{origin}}}{\text{Tokens}_{\text{origin}}} \times 100\%
\]

\xhdr{Baselines}
We compare DeepPrune against several competitive baselines: (1) \textbf{Sampling Methods:} cons@512 which samples 512 parallel traces with majority voting for self-consistency~\cite{wang2022self}, serves as our primary baseline for token reduction calculations. (2) \textbf{Confidence-based Pruning Methods:} DeepConf-high and DeepConf-low are confidence-based early stopping methods with high or low threshold for pruning. These baselines represent the state-of-the-art in efficient reasoning methods. We ensure fair comparison by using identical model checkpoints and experimental configurations with the DeepConf~\cite{fu2025deep}.

\begin{table*}[htbp]
\centering
\resizebox{\textwidth}{!}{%
\begin{tabular}{l|ccc|ccc|ccc}
\toprule
\multirow{2}{*}{\textbf{Metric}} & \multicolumn{3}{c|}{\textbf{DeepSeek-8B}} & \multicolumn{3}{c|}{\textbf{Qwen3-32B}} & \multicolumn{3}{c}{\textbf{GPT-OSS-20B}} \\
\cmidrule(lr){2-4} \cmidrule(lr){5-7} \cmidrule(lr){8-10}
 & \textbf{AIME24} & \textbf{AIME25} & \textbf{GPQA} & \textbf{AIME24} & \textbf{AIME25} & \textbf{GPQA} & \textbf{AIME24} & \textbf{AIME25} & \textbf{GPQA} \\
\midrule

cons@512${}^{\dag}$ & & & & & & & & & \\
\quad Token ($\times 10^8$) & 3.55 & 4.01 & 9.92 & 2.00 & 2.43 & 7.44 & 5.57 & 6.26 & - \\
% \quad $\Delta \text{Token}\%$ & - & - & - & - & - & - & - & - & - \\
\quad Accuracy & 86.7\% & 82.3\% & 72.5\% & 84.8\% & 80.1\% & 72.2\% & \underline{96.7\%} & 95.4\%& - \\
\midrule

DeepConf-high${}^{\dag}$ & & & & & & & & & \\
\quad Token ($\times 10^8$) & 1.45 & 2.37 & 6.90 & 0.88 & 1.61 & 4.16 & 3.07 & 3.18 & - \\
\quad $\Delta \text{Token}\%$ & -59.0\% & -40.9\% & -30.4\% & -56.0\% & -33.7\% & -44.1\% & -44.8\% & -49.2\% & - \\
\quad Accuracy & 86.7\% & 81.4\% & 72.4\% & 86.4\% & 80.2\% & 72.9\% & \underline{96.7\%} & 95.3\% & - \\
\midrule

DeepConf-low${}^{\dag}$ & & & & & & & & & \\
\quad Token ($\times 10^8$) & 0.78 & 1.24 & 3.46 & 0.66 & 1.14 & 3.21 & 1.11  & 1.21  & - \\
\quad $\Delta \text{Token}\%$ & -77.9\% & -69.0\% & -65.1\% & -66.8\% & -52.9\% & -56.9\% & \textbf{-80.0\%} & -80.7\% & - \\
\quad Accuracy & \underline{92.5\%} & \underline{86.4\%} & \underline{71.7\%} & 89.5\% & 80.2\% & \underline{73.0\%} & 95.7\% & \underline{96.1\%} & - \\
\midrule

cons@512 & & & & & & & & & \\
\quad Token ($\times 10^8$) & 3.62 & 4.19 & 10.9 & 1.93 & 2.64 & 6.94 & 2.05 & 2.10 & 4.60 \\
% \quad $\Delta \text{Token}\%$ & - & - & - & - & - & - & - & - & - \\
\quad Accuracy & 86.7\% & 83.3\% & 66.2\% & 86.7\% & 80.0\% & 70.7\% & 93.3\% & 90.0\% & \underline{70.7\%} \\
\midrule

\textbf{DeepPrune (ours)} & & & & & & & & & \\
% this is results for tau = 0.63
% \quad Token ($\times 10^8$) & 0.27 & 0.24 & 1.41 & 0.12 & 0.13 & 1.00 & 0.23 & 0.21 & 1.18 \\
% \quad $\Delta \text{Token}\%$ & \textbf{-92.7\%} & \textbf{-94.3\%} & \textbf{-87.1\%} & \textbf{-93.5\%} & \textbf{-95.1\%} & \textbf{-85.6\%} & \textbf{-89.1\%} & \textbf{-90.2\%} & \textbf{-74.6\%} \\
% \quad Accuracy & 86.7\% & 83.3\% & 61.6\% & 86.7\% & \underline{83.3\%} & 70.2\% & 93.3\% & 90.0\% & 66.7\% \\
% thi is results for tau = 0.5
\quad Token ($\times 10^8$) & \textbf{0.42} & \textbf{0.35} & \textbf{2.54} & \textbf{0.26} & \textbf{0.23} & - & \textbf{0.42} & \textbf{0.38} & \textbf{2.20} \\
\quad $\Delta \text{Token}\%$ & \textbf{-88.3\%} & \textbf{-91.6\%} & \textbf{-76.7\%} & \textbf{-86.4\%} & \textbf{-91.4\%} & \textbf{-\%} & -79.6\% & \textbf{-82.2\%} & \textbf{-52.5\%} \\
\quad Accuracy & 86.7\% & 83.3\% & 63.1\% & \underline{90.0\%} & \underline{90.0\%} & -\% & 90.0\% & 93.3\% & 68.7\% \\
\bottomrule
\end{tabular}%
}
\caption{Online experimental results showing token consumption (in $\times 10^8$) and accuracy across three reasoning models on three benchmarks. The table compares different methods including conventional sampling (cons@512), confidence-based approaches (DeepConf-high, DeepConf-low), and the proposed DeepPrune. Token savings relative to cons@512 ($\Delta \text{Token}\%$) are also provided where applicable. ${}^{\dag}$ indicates results taken from the DeepConf paper.}
\label{tab:transposed_results}
\end{table*}

\subsection{Offline Experiment Results}
\label{sec:offline_results} As shown in Table~\ref{tb:method_comparison}, we have three observations: (1) Our best configuration, which uses top-25 reasoning words with focal loss and oversampling, achieves superior performance with an average AUROC of 0.8701 and TNR@0.2 of 0.8186 across all models. This represents a substantial improvement over the preliminary zero-shot LLM judgment (AUROC=0.66) reported in Figure~\ref{fig:subfig3}, validating the necessity of specialized training for redundancy prediction. (2) The comparison between top-500 tokens and top-25 reasoning words reveals a clear advantage for reasoning-aligned truncation. This suggests that structural alignment with reasoning steps provides more reliable signals for predicting answer equivalence compared to fixed-length token windows. (3) The ablation study on training strategies demonstrates the critical importance of addressing class imbalance. The combination of focal loss and oversampling consistently delivers the best performance across both truncation types and all reasoning models. Notably, using oversampling alone significantly degrades performance (AUROC drops to 0.7610 for tokens and 0.7983 for reasoning words), indicating that simply balancing the dataset distribution is insufficient without proper loss weighting. Focal loss alone provides moderate improvements, but the synergistic combination with oversampling yields the most robust results.

% \xhdr{Cross-Model Generalization}
% The judge model exhibits strong generalization capabilities across diverse reasoning architectures. Particularly impressive is the performance on \texttt{GLM-4.5-Air}, where reasoning words with focal loss achieve the highest individual model AUROC of 0.8989. The consistent performance gains across all three test models---despite being trained exclusively on \texttt{Deepseek-R1-Distill-Llama-8b} traces---demonstrate that our approach learns transferable patterns of reasoning redundancy rather than model-specific artifacts.

% \xhdr{Implications for Online Pruning}
% The high TNR@0.2 scores (exceeding 0.81 for the best configuration) indicate that our judge model can effectively identify diverse reasoning paths while maintaining a low false negative rate. This property is crucial for online pruning, as it ensures that correct but diverse reasoning trajectories are preserved, thereby maintaining the performance benefits of parallel reasoning while achieving substantial computational savings.

\begin{table*}
\centering
\centering
\resizebox{\textwidth}{!}{
\begin{tabular}{l |c|c|c|c|c|c|c|c}
\toprule
% 列定义：左对齐（threshold） + 8个居中对齐列
% 第一层表头：模型名称
\multirow{3}{*}{Threshold} & \multicolumn{4}{c}{Qwen3-32B on AIME24} & \multicolumn{4}{c}{Qwen3-32B on AIME25} \\
% 第二层表头：聚类方法
& \multicolumn{2}{c}{Greedy Clustering} & \multicolumn{2}{c}{w. Majority Voting} & \multicolumn{2}{c}{Greedy Clustering} & \multicolumn{2}{c}{w. Majority Voting} \\
% 第三层表头：指标
 & Token($\times 10^8$) & pass@k & Token($\times 10^8$) & ACC & Token($\times 10^8$) & pass@k & Token($\times 10^8$) & ACC \\
\midrule
0.75 & 0.0148 & \underline{93.3\%} & 0.33 & 86.7\% & 0.0282 & \underline{96.7\%} & 0.31 & \underline{90.0\%} \\
0.63 & 0.0093 & \underline{93.3\%} & 0.28 & \underline{93.3\%} & 0.0265 & \underline{96.7\%} & \textbf{0.23} & 83.3\% \\
0.5 & 0.0082 & \underline{93.3\%} & 0.26 & 90.0\% & 0.0142 & 70\% & \textbf{0.23} & \underline{90.0\%} \\
0.25 & \textbf{0.0043} & \underline{93.3\%} & \textbf{0.25} & 90.0\% & \textbf{0.0050} & 70\% & \textbf{0.23} & \underline{90.0\%} \\
\bottomrule
\end{tabular}
}
\caption{Performance of DeepPrune with varying redundancy threshold $\tau$ on AIME datasets for Qwen3-32B. Token consumption, pass rate and accuracy are reported for two pruning settings: (1) Conduct greedy clustering then retains only one trace per cluster, (2) Perform majority voting to get one final answer with the largest cluster.}
\label{tb:ablation_3}
\end{table*}
\subsection{Online Experiment Results}
\label{sec:online_results}

Table~\ref{tab:transposed_results} presents the online evaluation results, from which we can get several key findings:

\xhdr{Substantial Token Reduction with Minimal Accuracy Loss}
DeepPrune achieves remarkable token savings while maintaining competitive accuracy across all experimental settings. Specifically, DeepPrune reduces token consumption by 52.5\% to 91.6\% compared to the cons@512 sampling baseline. The most significant reductions are observed on AIME datasets, where DeepPrune achieves 79.6\%-91.6\% token savings with negligible accuracy drop (within 3 percentage points). For instance, on Qwen3-32B with AIME25, DeepPrune reduces tokens by 91.4\% while even improving accuracy from 80.0\% to 90.0\%. This demonstrates that our method effectively identifies and eliminates redundant reasoning paths without compromising solution quality.

\xhdr{Superior Efficiency Compared to Confidence-Based Methods}
DeepPrune consistently outperforms confidence-based pruning methods: DeepConf-high and DeepConf-low, in terms of token efficiency. While DeepConf-low achieves substantial token reductions, DeepPrune provides the least token consumption across different configurations. More importantly, DeepPrune maintains more stable accuracy preservation compared to DeepConf-low, e.g., DeepPrune's 90.0\% vs DeepConf-low's 80.2\% on AIME25 with Qwen3-32B. This highlights the advantage of our inter-trace redundancy analysis over single-trace confidence estimation methods~\cite{fu2025deep}.

\xhdr{Cross-Model and Cross-Dataset Generalization}
The consistent performance across three distinct reasoning models (DeepSeek-8B, Qwen3-32B, and GPT-OSS-20B) and three diverse benchmarks (AIME24, AIME25, GPQA) validates the generalizability of our approach. Particularly noteworthy is that the judge model of DeepPrune was trained purely on the reasoning traces of \texttt{Deepseek-R1-Distill-Llama-8B}, which means all the tests are out-of-distribution, providing a practical solution for efficient parallel reasoning

% \xhdr{Optimal Performance on Mathematical Reasoning}
% DeepPrune demonstrates exceptional performance on mathematical reasoning tasks (AIME datasets), where it achieves the highest token reductions (89.1\%-95.1\%) with minimal or no accuracy degradation. This suggests that mathematical problems exhibit more predictable reasoning patterns that our judge model can effectively capture for early redundancy detection. The consistent performance across AIME24 and AIME25 further confirms the robustness of our approach to problem variations.

% The online results collectively demonstrate that DeepPrune successfully addresses the inter-trace redundancy problem identified in our preliminary analysis, providing a practical solution for efficient parallel reasoning while preserving answer diversity and quality.

\begin{figure*}[t]
    \centering
    \subfigure[Ablation of Token Number]{
        \includegraphics[width=0.46\linewidth]{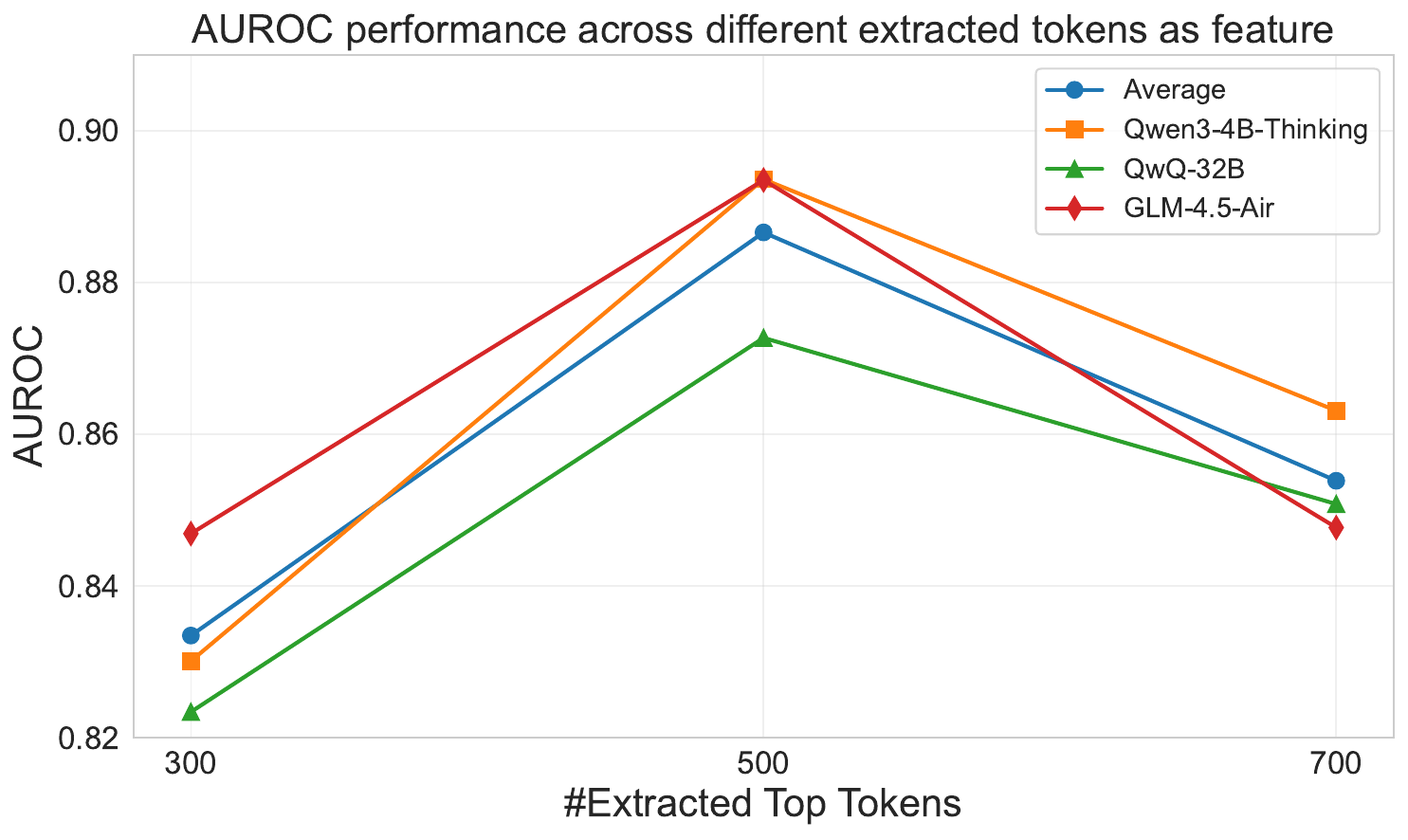}
        \label{fig:subfig_token}
    }
    \hfill
    \subfigure[Ablation of Reasoning Word Number]{
        \includegraphics[width=0.46\linewidth]{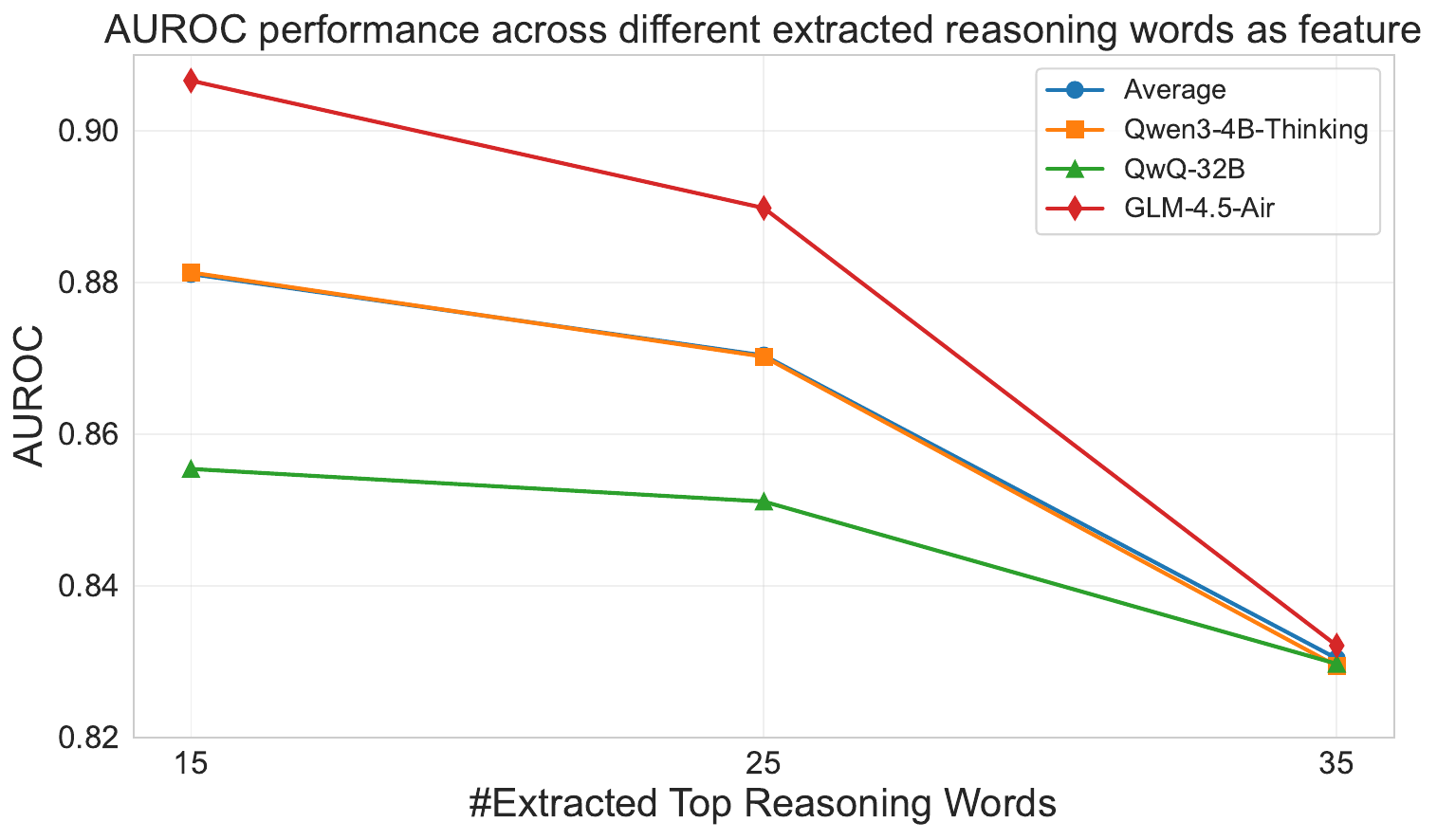}
        \label{fig:subfig_reasoning_word}
    }
 \caption{Ablation study on judge model with different truncation strategies for unfinished reasoning traces. We report the classification performance on three reasoning models' trace answer equivalence with different numbers of truncated top tokens (Figure (a)) and  different numbers of reasoning words in extracted segments (Figure (b)).}
    \label{fig:ablation_study}
\end{figure*}

% \begin{figure}[t]
%     \centering
%     \includegraphics[width=\linewidth]{figs/auroc_comparison_v1.pdf}
%     \caption{RAG performance across different context lengths, varied by including the top 4, 8, 16, 32, 64, 128, and 256 chunks of 512 tokens. The horizontal line show the overall score of each model without RAG at a full context length of 128k tokens.}
%     \label{fig:rag}
% \end{figure}

\subsection{Ablation Study}

To better understand the impact of different truncation strategies on the judge model's performance, we conduct an ablation study on the number of extracted top tokens and reasoning words. The results are presented in Figure~\ref{fig:ablation_study}. Comparing Figure~\ref{fig:subfig_token} and Figure~\ref{fig:subfig_reasoning_word}, it is evident that using reasoning words as features generally yields higher AUROC scores and a more pronounced optimal point compared to using raw top tokens. This confirms our hypothesis that extracting semantically rich reasoning words provides a more effective representation for the judge model, leading to better prediction of answer equivalence. The optimal performance is often found at an intermediate number of features (e.g., 500 tokens or 25 reasoning words), suggesting a sweet spot where sufficient context is provided without introducing excessive noise.

Besides, we analyze the trade-off between efficiency, answer diversity, and final accuracy by varying the redundancy threshold $\tau$ in DeepPrune (Table~\ref{tb:ablation_3}). The pass rate measures answer diversity after clustering, while accuracy reflects the voting outcome from the largest cluster.  As $\tau$ decreases from 0.75 to 0.25, token consumption decreases significantly due to more aggressive pruning. However, this comes at the cost of reduced answer diversity, particularly on challenging problems. On AIME25, pass rate drops from 96.7\% to 70\% under greedy clustering, indicating that higher thresholds may prune valuable diverse reasoning paths. The majority voting accuracy shows the same trend and the threshold $\tau=0.5$ provides the best balance.

%% file: 050conclusion.tex
\section{Conclusion}

We identify inter-trace redundancy as a major efficiency bottleneck in parallel reasoning, where over 80\% of computational resources are wasted on generating equivalent reasoning paths. To address this, we propose DeepPrune, a novel framework that trains a judge model on out-of-distribution data for online greedy clustering to dynamically prune redundant traces while preserving answer diversity. Extensive experiments show that our OOD-trained judge model generalizes strongly to unseen reasoning models, achieving 0.7072 AUROC in offline evaluation, and DeepPrune reduces token consumption by up to 88.5\% without hurting accuracy in online experiments. Our work establishes that learned similarity judgment effectively addresses redundancy in parallel scaling, paving the way for more efficient reasoning systems. 

%% file: 070limitation.tex
\section*{Limitations}
We acknowledge several limitations of our work. First, due to limited computational resources, our judge model is trained exclusively on reasoning traces from \texttt{Deepseek-R1-Distill-Llama-8B}, which may limit its generalization to other model families with distinct reasoning styles. While our cross-model evaluations show promising results, performance could potentially degrade on architectures different from the training distribution. Second, the greedy clustering algorithm, while efficient, makes locally optimal decisions that may occasionally prune beneficial diverse paths, particularly in complex reasoning scenarios where early similarity is not indicative of final answer equivalence. Third, our method introduces additional computational overhead from the judge model inferences during pruning. Although the overall token reduction is substantial, the relative efficiency gain depends on the cost ratio between the judge and reasoning models. Finally, the optimal redundancy threshold $\tau$ may be problem-dependent; while $\tau=0.5$ works well across our benchmarks, adaptive threshold selection could further improve performance. Addressing these limitations represents promising directions for future work.

\section*{Ethical Consideration}

We affirm that this work raises no significant ethical concerns. All models and datasets used in our experiments are publicly available with permissible licenses, ensuring proper attribution and compliant usage. Specifically, the reasoning models (DeepSeek-8B\footnote{\url{https://huggingface.co/deepseek-ai/DeepSeek-R1-0528-Qwen3-8B}}, Qwen~\cite{yang2025qwen3}, GPT-OSS~\cite{agarwal2025gpt}) and benchmarks (AIME~\cite{aime24,aime25}, GPQA~\cite{rein2024gpqa}) are widely recognized resources in the research community.

Our research focuses on improving the computational efficiency of reasoning processes through redundancy reduction, without involving sensitive data generation or manipulation. The content processed consists exclusively of mathematical and scientific reasoning tasks, which are devoid of personal, biased, or harmful material.

%% file: 080acknowledgement.tex
\section*{Acknowledgements}
This work is supported by the National Natural Science Foundation of China (625B2101, 62476150). This work is also supported by a grant from the Institute for Guo Qiang, Tsinghua University (2019GQB0003).

%% file: 060appendix.tex
\newpage
\appendix
\onecolumn

\section*{Appendix}

\section{Detailed Analysis of Inter-trace Redundancy}
\label{app:data_detail}

In Section~\ref{sec:pre_exp}, Figure~\ref{fig:subfig1} presents the distribution of redundant traces, showing the "same-answer" pair ratios for four different models. The average percentage of these models might not directly align with a simple arithmetic average of the individual percentages. This is because the amount of parallel reasoning trace data collected for each model varies significantly.

During the trace collection process, we initially sampled 16 responses for each query. However, some Chain-of-Thought traces failed to produce valid answers or did not terminate within the `max\_length` limit. These invalid traces were excluded from the pair calculation. Furthermore, while our full dataset comprised over 760 problems from benchmarks including GPQA, AIME24, AIME25, and Math500, due to computational resource constraints, we were only able to run all 760 queries on \texttt{Deepseek-R1-Distill-llama-8b}. For other models, we sampled approximately over 100 problems. This led to the following total pair counts and similarity ratios for each model in Table~\ref{tab:trace_collection_stats}.

\begin{table}[h!]
\centering
\begin{tabular}{l|c|c|c}
\toprule
\textbf{Model} & \textbf{Total Pairs} & \textbf{Same Answer Pairs} & \textbf{Similarity Ratio} \\
\midrule
GLM-4.5-Air & 11,870 & 11,213 & 0.9447 \\
QwQ-32B & 13,785 & 12,569 & 0.9118 \\
Deepseek-R1-Distill-llama-8b & 80,760 & 61,505 & 0.7616 \\
Qwen3-4B-Thinking-2507 & 13,800 & 12,852 & 0.9313 \\
Average & 120,215 & 98,139 & 0.8164 \\
\bottomrule
\end{tabular}
\caption{Detailed statistics of collected reasoning trace pairs and their similarity ratios across different models.}
\label{tab:trace_collection_stats}
\end{table}

Given the large disparity in the number of total pairs, particularly the substantial contribution from \texttt{Deepseek-R1-Distill-llama-8b}, a weighted average would be necessary to accurately reflect the overall inter-trace redundancy across the combined dataset. Our analysis in Section~\ref{sec:pre_exp} is based on the aggregate distribution, rather than a simple average of individual model ratios.

\section{Reconciliation of Online Experiment Results}

\subsection{Differences in cons@512 and DeepConf Baselines}
In Table \ref{tab:transposed_results} of our online experiments, the reported cons@512 and DeepConf baseline results might show slight differences compared to those originally published in the \texttt{DeepConf} paper. We conducted our experiments using the exact same three reasoning models (\texttt{DeepSeek-8B}, \texttt{Qwen3-32B}, and \texttt{GPT-OSS-20B}) and the same three benchmarks (AIME 2024, AIME 2025, and GPQA) as in the \texttt{DeepConf} paper.

The cons@512 method, which involves sampling 512 responses, inherently has a degree of randomness. To minimize these discrepancies and ensure a fair comparison, we meticulously aligned our experimental setup with that described in the \texttt{DeepConf} codebase. This included using identical sampling temperatures and `max\_length` settings for each model's hyperparameters, and employing vllm~\cite{kwon2023efficient} as the inference engine.

Despite these efforts, minor differences in the final evaluation results persisted. To further align our results, we initially sampled 640 responses (512 + 128) and then re-sampled 512 responses. This re-sampling process was performed to match the pass@1 metrics for each model and dataset as closely as possible to those reported in the original \texttt{DeepConf} paper, ensuring the differences were within 3 percentage points. The aligned pass@1 values are presented in Table~\ref{tab:pass_at_1_alignment}.

\begin{table}[h!]
\centering
\begin{tabular}{l|l|c|c}
\toprule
\textbf{Model} & \textbf{Dataset} & \textbf{DeepPrune Pass@1} & \textbf{DeepConf Pass@1} \\
\midrule
Qwen3-32B & AIME\_2025 & 69.28 & 71.7 \\
Qwen3-32B & AIME\_2024 & 80.10 & 80.6\\
Qwen3-32B & GPQA & 68.24 & 68.9 \\
DeepSeek-R1-0528-Qwen3-8B & AIME\_2025 & 74.57 & 76.9 \\
DeepSeek-R1-0528-Qwen3-8B & AIME\_2024 & 83.08 & 83.0 \\
DeepSeek-R1-0528-Qwen3-8B & GPQA & 59.74 & 62.8 \\
GPT-OSS-20B & AIME\_2025 & 77.54 & - \\
GPT-OSS-20B & AIME\_2024 & 80.72 & - \\
GPT-OSS-20B & GPQA & 66.18 & - \\
\bottomrule
\end{tabular}
\caption{Comparison of pass@1 metrics with the same values taken from the DeepConf paper after alignment procedure. Differences are within 3 percentage points.}
\label{tab:pass_at_1_alignment}
\end{table}
This careful alignment allows for a more direct comparison of DeepPrune's performance against established baselines.

\subsection{Missing DeepConf Result for GPT-OSS-20B on GPQA}
In Table \ref{tab:transposed_results}, the entry for DeepConf on GPT-OSS-20B for the GPQA dataset is marked as empty (-). This is because the original \texttt{DeepConf} paper did not report experimental results for this specific model-dataset combination. Due to our limited computational resources, we were unable to conduct this additional experiment to fill the gap.

\subsection{Dataset and Model Specifications}
To ensure full transparency and reproducibility, we explicitly state the versions and sources of the datasets and models used in our experiments. If not stated below, then the name of the dataset or model should be its full name.
\begin{itemize}
    \item All references to \textbf{GPQA} in this paper, including those in our online experiments (Table~\ref{tab:transposed_results}) and trace collection, refer specifically to the \textbf{GPQA Diamond} subset, which consists of 198 problems. This is consistent with the dataset used in the \texttt{DeepConf} paper and can be accessed at \url{https://huggingface.co/datasets/fingertap/GPQA-Diamond}.
    \item The model referred to as \textbf{DeepSeek-8B} throughout this paper, corresponds to the \texttt{DeepSeek-R1-0528-Qwen3-8B} model. This is the same model variant employed in the \texttt{DeepConf} paper and is publicly available at \url{https://huggingface.co/deepseek-ai/DeepSeek-R1-0528-Qwen3-8B}. And \textbf{Qwen3-4B-Thinking} refers to  \url{https://modelscope.cn/models/Qwen/Qwen3-4B-Thinking-2507}.
\end{itemize}

\section{Hyperparameters}
Table \ref{tab:hyperparameters} summarizes the key hyperparameters used in the DeepPrune framework and their default values, along with a brief explanation of their meaning.

\begin{table}[h!]
\centering
\begin{tabular}{c|c|l}
\toprule
\textbf{Symbol} & \textbf{Default Value} & \textbf{Description} \\
\midrule
$k$ & 500 & Number of tokens for fixed-length prefix truncation \\
$k$ & 25 & Number of reasoning words for aligned segment truncation \\
$\gamma$ & 2.0 & Focal loss focusing parameter \\
$\alpha_t$ & 0.25  & Focal loss class-balancing parameter \\
$\tau$ & 0.5 & Similarity threshold for greedy clustering \\
$K$ & 16 & Maximum number of clusters \\
$p$ & 20 & Number of sampled traces for cluster similarity calculation \\
$q_1$ & 20 & Max traces to finish reasoning from the largest cluster \\
$q_2$ & 32 & Number of traces to sample for reasoning if all clusters are singletons \\
\bottomrule
\end{tabular}
\caption{Key hyperparameters of the DeepPrune framework. There are two $k$ symbols because each of them will not exist when the other is used since we can only use one truncation strategy.}
\label{tab:hyperparameters}
\end{table}

\section{Computational Resources}
The majority of our experiments, including the training of the judge model and a significant portion of the testing, were conducted on 4 NVIDIA A100 GPUs with 80GB memory each. In the later stages of the experiments, we also utilized a server equipped with 8 H20 GPUs for approximately two days. Our computational resources were relatively limited, especially considering the requirement to generate 512 responses from large language models for baseline comparisons.

\section{Discussion on Answer Diversity and Majority Voting}

\subsection{Comparison of Diversity with Baselines}
In the introduction, we highlight that a limitation of confidence-based early stopping is its potential impact on answer diversity, which DeepPrune aims to preserve. However, in our online experiments (Table~\ref{tab:transposed_results}), we primarily compare against baselines using accuracy rather than an explicit diversity metric like pass@k.

In reasoning tasks, answer diversity is often quantified by pass@k, which measures the proportion of samples where at least one of the top-$k$ retained answers is correct. Prior methods like \texttt{DeepConf} and cons@512 typically aggregate multiple traces to produce a single final answer, effectively reducing $k$ to 1 for their final output. Therefore, a direct pass@k comparison with these methods would be unfair, as our approach inherently produces a set of diverse reasoning paths (clusters) from which multiple candidates could be drawn, leading to a $k$ value greater than 1.

Our approach, after clustering, theoretically retains distinct reasoning paths in different clusters. Each cluster's representative trace could contribute to a pass@k calculation, where $k$ would typically be greater than one (equal to the number of clusters). This inherent diversity in our retained traces is evident in Table \ref{tb:ablation_3}, where higher $\tau$ thresholds lead to higher pass rates under greedy clustering, indicating more diverse paths are preserved. To ensure a fair comparison on a single-answer basis, our main online evaluation focuses on the final accuracy obtained after majority voting, as this is the metric that DeepConf and cons@512 also optimize for.

\subsection{Analysis of Low Pass Rate with Low Thresholds}
A notable observation in Table~\ref{tb:ablation_3} is that for Qwen3-32B on AIME25, the pass@k for "Greedy Clustering" at lower redundancy thresholds (0.5 and 0.25) is 70.0\%, which appears lower than the corresponding "w. Majority Voting" setting's accuracy. This seemingly counter-intuitive result can occur because, at very low similarity thresholds, the judge model becomes highly permissive, predicting a large number of trace pairs as having the same answer.

This permissiveness causes many traces to be grouped into a few large clusters, leading to an uneven distribution of traces across clusters. When we calculate the pass@$k$ metric for "Greedy Clustering" we typically select one representative trace from each unique cluster to contribute to the diversity count (i.e., $k$ equals the number of distinct clusters). If most traces are concentrated in only a few clusters, the effective $k$ becomes very small. In such scenarios, even if the individual clusters contain correct answers, the limited number of distinct clusters (and thus $k$ value) can result in a lower pass@k percentage, as it does not fully capture the potential for correctness from the overall set of generated traces.

Conversely, for the "w. Majority Voting" results, our strategy in these low-threshold scenarios is to sample up to $q_1=20$ traces from the largest cluster to perform majority voting (as described in Section \ref{sec:majority_voting}). This allows us to leverage a greater number of individual traces to reach a consensus, ensuring a more robust final answer and potentially leading to higher accuracy, even when the overall number of distinct clusters and pass@k is low. This mechanism helps maintain effective accuracy by focusing computational resources on the most prominent reasoning paths, despite a reduced perceived diversity at the cluster level when our judge model dose not act perfectly.
\newpage

\subsection{Rationale for Majority Voting}
We employ majority voting in DeepPrune for two primary reasons:
\begin{enumerate}
    \item \textbf{Effectiveness and Common Practice:} Majority voting is a widely adopted and empirically effective method for aggregating multiple reasoning traces to derive a robust final answer, as demonstrated by pioneering works like Self-Consistency~\cite{wang2022self}. It helps mitigate errors from individual traces and leverages the collective intelligence of diverse reasoning paths.
    \item \textbf{Fair Comparison with Baselines:} To enable a direct and fair comparison of final answer accuracy with methods like DeepConf that also produce a single aggregated answer, we needed a mechanism to consolidate the diverse traces retained by DeepPrune into one final prediction. While our method inherently preserves inter-trace diversity for potential pass@k evaluations (as explored in Table \ref{tb:ablation_3}), majority voting allows us to align with the single-answer output paradigm of competitive baselines.
\end{enumerate}

It is important to note that DeepPrune is highly flexible and can be integrated with other aggregation strategies. For instance, instead of simple majority voting, one could employ a selection model~\cite{moshkov2025aimo} to choose the best answer from the diverse set of clusters. Furthermore, DeepPrune is orthogonal to existing method; for example, after our method reduces inter-trace redundancy, a confidence-based filter like DeepConf could be applied within each remaining cluster or across selected representatives to further refine the final answer. Due to our limited resources, we leave the exploration of these alternative aggregation strategies and combinations for future work.